\documentclass[letterpaper, 10 pt, conference]{ieeeconf}

\IEEEoverridecommandlockouts                              

\usepackage{graphics} 
\usepackage{epsfig} 

\usepackage{times} 
\usepackage{amsmath} 
\usepackage{amssymb}  

\usepackage{todonotes}
\usepackage{caption}
\usepackage{subcaption}
\usepackage{hyperref}
\makeatother
\title{\LARGE \bf
Learning Pregrasp Manipulation of Objects from
Ungraspable Poses
}

\author{Zhaole Sun$^{1}$, Kai Yuan$^{2}$, Wenbin Hu$^{2}$, Chuanyu Yang$^{2}$, and Zhibin Li$^{2}$
    \thanks{This work was supported by the National Natural Science Foundation  of China under Grant 61876054, the China Scholarship Council, the EPSRC CDT in Robotics and Autonomous Systems (EP/L016834/1), EPSRC Future AI and Robotics for Space (EP/R026092/1), and Offshore Robotics for Certification of Assets (EP/R026173/1)..}
    \thanks{$^{1}$Zhaole Sun is with the Department of Automation, Tsinghua University, China
        {\tt\small szl16@mails.tsinghua.edu.cn}}%
    \thanks{$^{2}$Kai Yuan, Wenbin Hu, Chuanyu Yang, and Zhibin Li are with the School of Informatics, University of Edinburgh, UK
    }
}

\begin{document}
\bstctlcite{IEEEexample:BSTcontrol}
\maketitle
\thispagestyle{empty}
\pagestyle{empty}

\begin{abstract}
In robotic grasping, objects are often occluded in ungraspable configurations such that no pregrasp pose can be found, eg large flat boxes on the table that can only be grasped from the side. Inspired by humans' bimanual manipulation, eg one hand to lift up things and the other to grasp, we address this type of problems by introducing pregrasp manipulation -- push and lift actions. We propose a model-free Deep Reinforcement Learning framework to train control policies that utilize visual information and proprioceptive states of the robot to autonomously discover robust pregrasp manipulation. The robot arm learns to first push the object towards a support surface and establishes a pivot to lift up one side of the object, thus creating a clearance between the object and the table for possible grasping solutions. Furthermore, we show the effectiveness of our proposed learning framework in training robust pregrasp policies that can directly transfer from simulation to real hardware through suitable design of training procedures, state, and action space.
Lastly, we evaluate the effectiveness and the generalisation ability of the learned policies in real-world experiments, and demonstrate pregrasp manipulation of objects with various size, shape, weight, and surface friction.
\end{abstract}

\section{Introduction}

Grasping is one of the most fundamental aspects of robotic manipulation. Previous works concentrate on grasp quality evaluation \cite{roa2015grasp,mahler2017dex} and grasp detection \cite{redmon2015real,kumra2017robotic, lenz2015deep} to predict a pose for grasping. Leveraging techniques of supervised learning, methods such as Dex-Net \cite{mahler2017dex} achieved a grasping accuracy of more than 90\% and exhibited a grasp efficiency comparable to human performance, even in clutter as being crowded with many other irrelevant objects.

Noticeably, objects in the aforementioned settings usually have several graspable positions, and the proposed methods \cite{mahler2017dex, mahler2019learning} use grasp quality evaluation in a pipeline as: Sample several feasible grasping positions on the object, rank them and then select the best position to grasp.
However, in scenarios where there is no feasible position to grasp, those methods would fail due to the non-existence of feasible grasping solutions in such particular configurations. 

\begin{figure}[t]
	\centering
	\includegraphics[width=80mm]{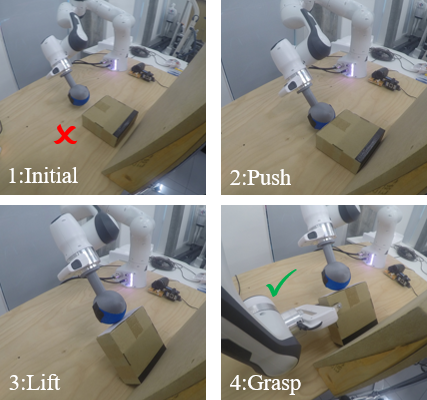}
	\caption{Demonstration of our transferred sim2real pregrasp policy. As the width of the box exceeds the gripper's size, the policy needs to learn a feasible grasping pose by a specific sequence of actions: (1) push the object against a support surface; (2) lift the object by pivoting; (3) grasp the object.}
	\label{fig:firstPageDemo}
	\vspace{-8mm}
\end{figure}

A notable scenario is that a cuboid with only its height being less than the max stroke of the gripper laying on flat ground (see Fig. \ref{fig:firstPageDemo}). In this configuration, no feasible grasp solution exists, and the object needs to be lifted first creating a feasible clearance, before being possibly grasped. Hereby, we name \textit{pregrasp manipulation} as the manipulation generating graspable configurations for finding suitable grasping solutions.

Recently, pregrasp sliding \cite{hang2019pre} and pregrasp rotation \cite{chang2008preparatory} has been proposed to tackle the pregrasp problem. Pregrasp sliding concentrates on creating grasping positions by sliding thin objects over the edge of the table, while pregrasp rotation focuses on rotating sensitive objects, such as rotating a pan's handle to improve grasp performance. These algorithms have strict conditions on the problem formulation and experimental setup, such as carefully designed end-effectors, precise and expensive sensors, a more complicated measure on the environment, and assumptions on the target object parameters. Also, pregrasp sliding \cite{hang2019pre} needs to have the information of the exact shape of the table, and a force-torque sensor was also used in the work of Hou et al. \cite{hou2019robust}. Thus, these methods have limitations to directly solve the problems in the setting of large objects with unknown structures, material composition, and uncertain weight and stiffness.

Since once a feasible grasping pose is established, the state of the art grasping algorithms are likely to be able to perform successful grasping, we hence focus on the study to create feasible, pregrasp positions for the existing state of the art algorithms. In contrast to other existing pregrasp methods, we do not rely on additional environmental objects and require no assumptions on the shape or type of object for pregrasp. This framework has potential to be integrated later as a pipeline together with existing grasping algorithms.  

As experiments showed, our proposed method is robust to environmental variations as we can use a second manipulator or a fixed support surface, against which the robot can successfully push and lift up the object for grasping. We thus do not require a suitable table edge as setup for the block to expose its graspable place (in contrast to pregrasp sliding \cite{hang2019pre}) and pose no assumptions on the shape of the object (in contrary to pregrasp rotation \cite{chang2010planning}).

In this work, we propose a Deep Reinforcement Learning Framework for pregrasping and apply the learned policy in real-world scenarios (Fig. \ref{fig:firstPageDemo}). In particular, we train a pregrasp policy in simulation through model-free learning-based continuous control. To demonstrate the effectiveness of the proposed design of the learning framework, we directly transfer the policy trained in simulation to reality (sim2real). 

The main contributions are as follows:
\begin{itemize}
\item A Deep Reinforcement Learning Framework for pregrasp manipulation to generate feasible grasping poses.
\item Proposal of environment-independent state and reward representations, allowing direct sim2real transfer of the trained pregrasp policy.
\item Robust and generalised ability of the trained pregrasp policy on evaluation metrics.
\end{itemize}
In the following sections, we first review related work in Section II, and present our proposed learning framework in Section III. In Section IV, we describe our experiment settings and methods for direct sim2real transfer. All real-world experiment results are analysed, compared and presented in Section V. Lastly, we conclude our work in Section VI.

\begin{table*}[t]
    \centering
    \caption{Comparison across related algorithms on each aspect for pregrasp manipulation or lifting a thin object. }
    \begin{tabular}{lllllll}
        \hline
        Work & End Effector & Target & Manipulation & External Condition & Method & Perception \\ \hline
        Proposed & Sphere & Rigid objects & Push\&Lift & Fixed object & DRL & RGB Camera \\
        Hang et al. \cite{hang2019pre}   & Fingertip & Thin Objects & Slide & Table Edge & Tree Search & Table Modeling \\
        Babin et al. \cite{babin2018picking}  & Gripper & Small or Thin Objects & Nail-Scoop &  & Quasistatic Mechanism & Known Object \\
        King et al. \cite{king2013pregrasp}& Hand & Normal Objects & Slide &  & Gradient Optimisation & Known Object \\
        Hou et al. \cite{hou2019robust} & Rubber Ball & Cuboid & Tilt(Lift) & Vertical Wall & Force-Velocity Control & FT Sensor \\
        Chang et al. \cite{chang2008preparatory}& Hand & Rotation-Sensitive Objects & Rotation & Motion Data & Human Demonstration & Vicon camera system \\ \hline
    \end{tabular}
    \label{tbl:compare_works}
    \vspace{-5mm}
\end{table*}
\section{Related Work}

\subsection{Pregrasp Manipulation}
Grasping is an active research topic in robotics. Many algorithms have been proposed to give improved grasp planning, policy, or detection \cite{mahler2017dex,morrison2018closing,liang2019pointnetgpd}. But these works all have a common assumption that the object can be grasped directly from at least one suitable position. In some scenarios, this assumption is not true, for example using a small gripper to grasp large flat objects. Pregrasp manipulation is introduced to solve some grasp tasks under such constrained conditions. Its intuition lays on using additional manipulation before grasping to improve grasp quality. Chang et al. proposed automatic planning of pregrasp rotation for object transport tasks \cite{chang2010planning}. Recently, Hang et al. use pregrasp sliding to move thin objects by re-configured hands to the edge of the table for grasping \cite{hang2019pre}. To deal with grasping an object in cluttered and uncertain environments, Moll et al. raise a rearrangement method to make space for the gripper to reach and grasp the target \cite{moll2017randomized}. 

\subsection{Deep Reinforcement Learning for Grasping}
Deep Reinforcement Learning (DRL) has achieved great success with the development of human-level game policies on strategy games \cite{silver2017mastering} and video games \cite{mnih2013playing}. Due to the success of implementing DRL to solve control problems in physical simulation environments like Pybullet \cite{coumans2016pybullet} and Mujoco \cite{todorov2012mujoco}, the idea of using DRL to solve robotic manipulation problems has also become increasingly popular. Zeng et al. propose methods based on DQN to evaluate the highest probability grasp position in a clutter \cite{zeng2018robotic,zeng2018learning}. Levine et al. use policy search method to train a end-to-end policy for robotic manipulation \cite{levine2016end}. 
Quillen et al. \cite{quillen2018deep} provide a comparative evaluation of multiple DRL methods on vision-based robotic grasping, including Q-learning \cite{mnih2013playing}, deep deterministic policy gradient (DDPG) \cite{lillicrap2015continuous}, Monte Carlo policy evaluation \cite{sutton2018reinforcement} and so on. Compared to traditional control based algorithms, DRL is better at handling high-dimension problems with larger action space while requiring less human prior knowledge \cite{andrychowicz2018learning}.

\subsection{Comparison with Previous Works}
To distinguish our method from several related methods we list the key differences between our work and previous work in Table \ref{tbl:compare_works}. We compared our work against previous work that either focused on pregrasp manipulation to enhance grasp or grasping thin and flat objects on the table. This comparison covers several aspects:

\textit{1. Target objects:} Some works pose assumption on task-dependent objects to be pregrasped. We chose to focus on pregrasp rigid objects and pose no assumptions on the shape or material of the rigid object.

\textit{2. Choice of end effector:} Reconfiguring the common parallel-gripper end effector to better improve grasp or pregrasp manipulation has been a popular choice amongst various works:
    (1) Soft, compliant, or under-actuated end-effector to keep contact with the object while sliding or rotating the object for pregrasping \cite{hang2019pre}. 
    (2) 24-Degree of Freedom (DoF) Shadow Hand C3 end-effector to rotate objects for better grasping from human demonstrations \cite{chang2008preparatory}.
    (3) Rubber ball as 2 DoF end-effector to lift or tilt objects with large friction \cite{hou2019robust}.     
    (4) Two-fingertip gripper as end-effector \cite{babin2018picking}. One fingertip serves as a support surface for the object to attach to, while the other sharper fingertip mimicking a thumb slides beneath the thin target object.
    (5) Three-finger gripper as end-effector \cite{king2013pregrasp}. For a pregrasping, all three fingers are all moved in one direction forming an open palm to push the target object into a better grasp configuration.
    In this work, we chose a spherical end-effector due to the increased surface contact with the target object, and speed of collision calculations in simulation.

\textit{3. External Condition:} To perform pregrasp all methods assume one external condition for pregrasp manipulation. For our method, we require the existence of a support surface that is a static and fixed object, e.g., a wall, against which the target object can be pushed against for lifting.

\textit{4. Manipulation:} They type of motion enabling the pregrasp configuration. We first push the target object against a static object and then lift by leveraging the point of contact between the target and fixed object as a pivot point.

\textit{5. Method:} Describes the core methods to be generated the pregrasp motion. In our work, a DRL agent is able to learn a pregrasp policy by autonomously exploring and generating training data from which it infers reward signals to accomplish the task.

\textit{6. Perception:} For obtaining the target object state, perception is required. The required sensors vary in their precision, cost, and capability. Generally, the more sophisticated the sensing mechanism, the fewer assumptions are posed on the experimental setup. Our work strikes a good balance between cheap and expressive external sensing (RGB camera) and generality of the target object (any rigid object).

To pregrasp the target object, our method uses a spherical end-effector and requires a fixed support surface to push the target object against. In Section V we show that the support surface can be a variety of surfaces, such as a wall, the gripper of a robot arm, objects with unusual form, and deformable objects. Furthermore, through training in our proposed DRL framework, the policy exhibits robustness and generalisation as shown in Section V.

\section{Training a Pregrasp Policy via DRL}
In this section, we will present a learning framework for pregrasp manipulation. In particular, we leverage the sample-efficiency and exploration characteristics of the Soft-Actor-Critic (SAC) algorithm \cite{haarnoja2018soft} to train a policy to solve this task. We further detail all necessary design choices and training procedures for the pregrasp task to enable a direct transfer of the policy trained in simulation to the real robot.

\subsection{Policy Optimization}
\begin{figure*}[t]
    \centering
    \includegraphics[width=\textwidth]{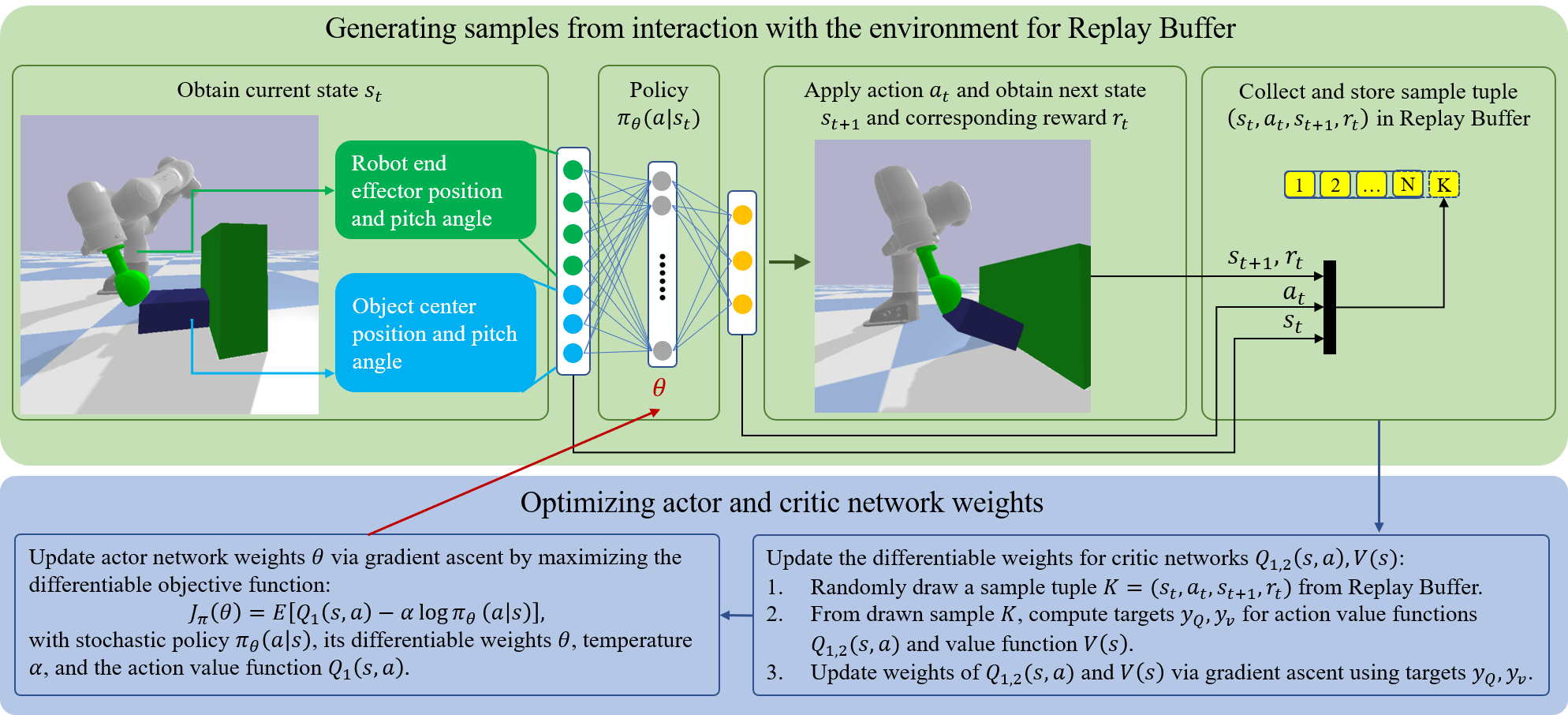}
    \caption{An overview of the proposed pipeline of pregrasp manipulation.}
    \label{fig:lifter_pipeline}
\end{figure*}
The overall learning pipeline in how to obtain a robust pregrasp policy $\pi_\theta(a|s_t)$ is shown in Fig. \ref{fig:lifter_pipeline}. 
We model the environment as an infinite horizon Markov Decision Process (MDP). Every state transition within this MDP can be defined by a tuple $(s_t,a_t,r_t,s_{t+1})$ consisting of the state $s_t$, action $a_t$ in the continuous action space $\mathcal{A}$, the resulting state $s_{t+1}$, and the reward $r_t$ returned by the environment.
For policy, at any given state $s_t$ at time t, the agent get an action $a_t$ according to learned policy $\pi(s_t)$, and we denote the state and state-action marginals and trajectory distribution $\rho_\pi(s_t)$ and $\rho_\pi(s_t,a_t)$ induced by this policy following original SAC denotations and settings \cite{haarnoja2018soft}.

For SAC, the policy $\pi$ aims to maximize the expected cumulative soft value objective: 
\[ J(\pi) = \sum_{t=0}^{T} E_{(s_t,a_t)\sim \rho_\pi} [r(s_t,a_t) + \alpha H(\pi(\cdot|s_t))], \]
with temperature $\alpha$ for the policy entropy $H(\pi(\cdot|s_t))$, which encourages agent exploration when being enlarged. 

During training and exploration, the stochastic action is being sampled from a Gaussian policy $\pi(s) = \mu(s) + \sigma(s)$ with deterministic policy $\mu(s)$ and standard deviations $\sigma(s)$. Both deterministic policy, and standard deviation are parametrized as 2 layered Multi-Layer Perceptron (MLP) with 64 neurons each using a tanh activation function.

\subsection{State Representation}
\begin{figure}[t]
\vspace{-3mm}
    \centering
    \includegraphics[width=0.48\textwidth]{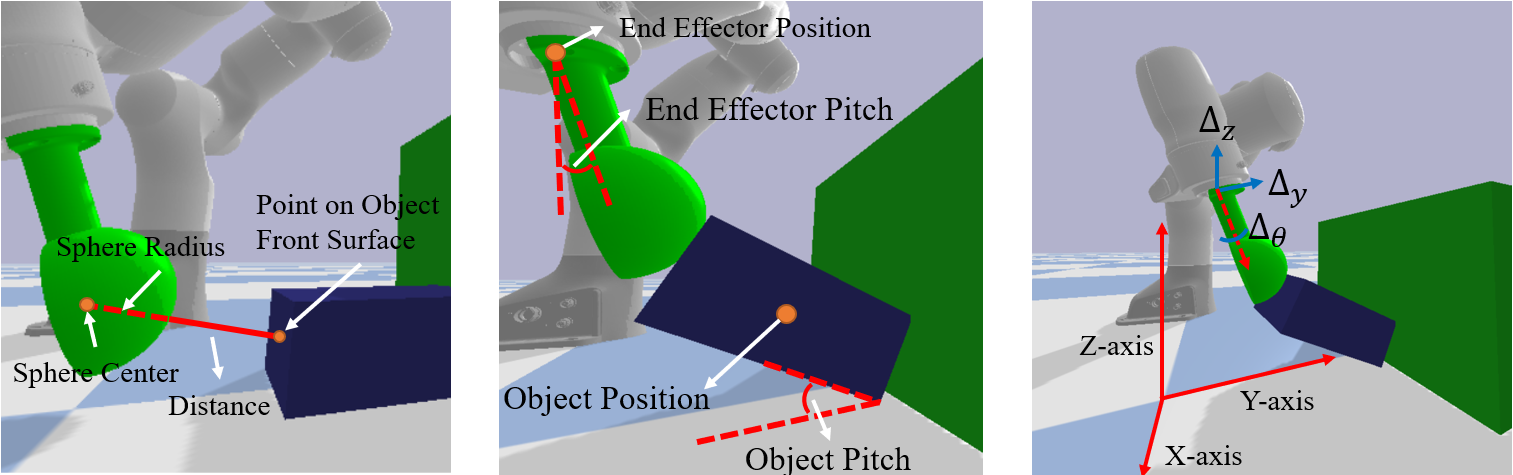}
    \caption{State and action space for the pregrasping policy.
        Left: calculation of distance $d$ between the end-effector and the object. Middle: definition of the end-effector and target object poses. Right: action space in global coordinates.}
    \label{fig:actionState}
\vspace{-5mm}
\end{figure}
The state input (Fig. \ref{fig:actionState} left and middle) consists of proprioceptive sensor information of the robot and perception information about the manipulated object from an external camera. The state $s_t$ is defined as a 7-dimension vector: 
\begin{equation}
s = [ d, p_{\text{eff}, y},  p_{\text{eff}, z}, \theta_{\text{eff}}, p_{\text{target}, y}, p_{\text{target}, z}, \theta_{\text{target}}],
\end{equation}
with distance $d$ between the end effector's sphere surface to the front surface of the target object, centre positions $p_{\text{target}, y}, p_{\text{target}, z}, p_{\text{eff}, y},  p_{\text{eff}, z}$, and pitch orientations $\theta_{\text{target}}, \theta_{\text{eff}}$ of the target object and end effector respectively.

The distance $d \geq 0$ between end-effector and the front surface of the target object is normalized equating to zero during contact.
The centre of the spherical end-effector are obtained via Forward Kinematics. The centre of the front surface of the object is extracted from the visual information.

\subsection{Action Representation}
The policy $\pi(s_t)$ outputs an incremental action:
\begin{equation}
a = [\Delta_\text{y}, \Delta_\text{z}, \Delta_\text{pitch}]
\end{equation}
where the action bounds of $\Delta_y$ and $\Delta_z$ range between $-0.025m \leq \Delta_{y,z} \leq 0.025m$. The action bound for the pitch orientation is $-0.01 \leq \Delta_{\text{pitch}} \leq 0.01$.

The policy network's output $\text{action}$ is incrementally added to the current end effector target command:
\begin{equation}
\begin{bmatrix}
p_{\text{eff}, y}\\
p_{\text{eff}, z}\\
\theta_{\text{eef}}
\end{bmatrix}_{t+1}=
\begin{bmatrix}
p_{\text{eff}, y}\\
p_{\text{eff}, z}\\
\theta_{\text{eef}}
\end{bmatrix}_{t} + 
\begin{bmatrix}
\Delta_\text{y}\\
\Delta_\text{z}\\
\Delta_\text{pitch}
\end{bmatrix}_{t}
\end{equation}

The target end-effector pose of the robot is further limited to be within the range of the dexterous workspace, and without colliding with the static supporting surface (e.g., wall). We terminate the episode if undesired collision occurs. To keep the object from continually rotating, we set a maximum angle 0.785 rad which is around $45^{\circ}$ in training, and the agent will also terminate the episode when the pitch angle is larger than 0.785 rad.

\subsection{Reward Design}
We follow a paradigm of simple reward design that is independent from the reality gap between simulation and reality, while not over-constraining the emerging motions through over-engineering the reward. To prevent reward exploitation, we regularly validate the sub-reward terms in their correctness. We found that the following reward is able to robustly pregrasp any rigid target object:
\begin{align}
 r = \lambda_1 r_{pitch} + \lambda_2 r_{dist},~~
 r_{pitch} = \theta_{\text{target}},~~r_{dist} = -d
\end{align}
with positive reward weights $\lambda_{1}, \lambda_2$, negative distance reward $r_{dist}$ for contact between end-effector and target object while pushing, and target pitch orientation reward $r_{pitch}$ for lifting the object as high as possible.

\subsection{Increasing Policy Robustness for sim2real}
\label{sec:sim2real}
To train a robust pregrasp policy that is able to reliably act under uncertainties, such as the reality gap between simulation and reality and inherent noise in the action and state space, we train the policy in randomly changing environments. 
During the initialisation stage of the simulation environment, we randomize the quantities shown in Table \ref{tbl:initialision}. 

Further, to bias the sample distribution towards high reward success states, while omitting infeasible states, we initialise the robot during training in multiple reference states yielding high results, or where it struggles to find a solution. Furthermore, we terminate the episode in undesired states, such as self-collision, or being outside of the workspace.

\begin{table}[t]
\vspace{-3mm}
\caption{Randomisation of physics parameters for training a robust pregrasp policy. All default values are uniformly randomized within their bounds.}
\centering
\begin{tabular}{lcccc}
\hline
Parameters Initialisation & Default & Min & Max & Probability \\ \hline
Mass(kg) & 0.08 & 0.02 & 0.10 & 0.30 \\
Friction Coefficient & 0.40 & 0.20 & 0.80 & 0.25 \\
Object Position(m) & 0.21 & 0.16 & 0.23 & 0.20 \\
Support Object Position(m) & 0.35 & 0.32 & 0.38 & 0.05 \\
$p_{\text{eff}, y}$(m) & 0.00 & -0.30 & 0.10 & 0.40 \\
$p_{\text{eff}, z}$(m) & 0.18 & 0.17 & 0.25 & 0.40 \\
$\theta_{\text{eef}}$(rad) & -2.75 & -2.40 & -2.80 & 0.40 \\ \hline
Reference Initialisation & \multicolumn{1}{l}{} & \multicolumn{1}{l}{} & \multicolumn{1}{l|}{} & 0.05 \\ \hline
\end{tabular}
\label{tbl:initialision}
\vspace{-5mm}
\end{table}

\section{Experiment Setting}
\begin{figure}[h]
    \centering
    \includegraphics[width=0.48\textwidth]{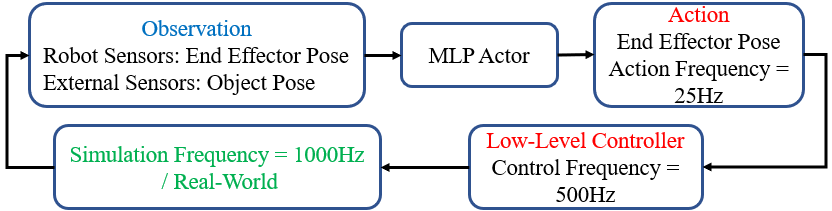}
    \caption{Control diagram for both simulation and real tests: states estimation (1kHz), actions (25Hz), control frequency of the robot arm (500Hz). The physics is simulated at 1kHz.}
    \label{fig:controlDiagram}
    \vspace{-5mm}

\end{figure}

This section presents the experimental setup for simulation and real experiments, the metrics for evaluating the trained pregrasp policy, and the overall control diagram (Fig. \ref{fig:controlDiagram}).

\subsection{Simulation Setup}
\begin{figure}[t]
    \centering
    \includegraphics[width=0.48\textwidth]{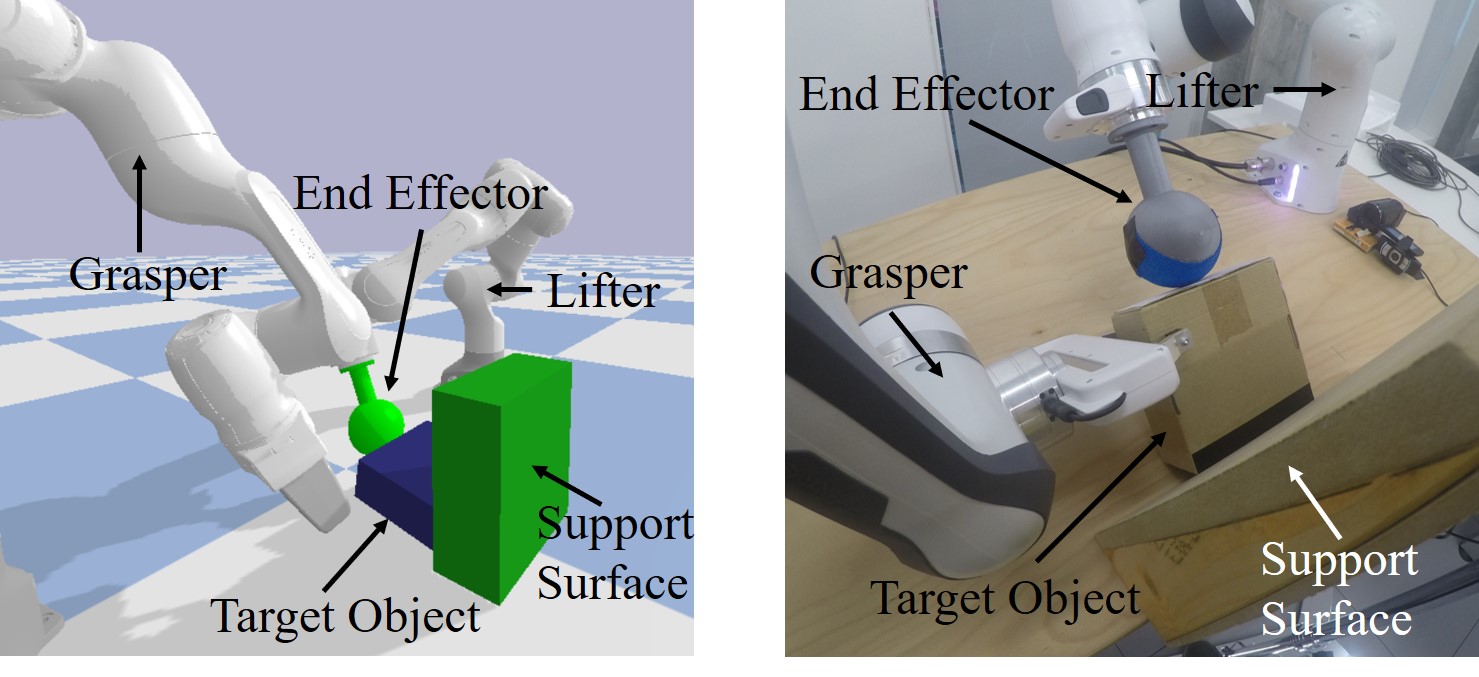}
    \caption{Experiment setup. {Left: }Simulation setup in Pybullet. {Right: }Real-world setup. The base of the grasper is 1 meter away from the base of the lifter. The target and support objects are placed between the two robot arms.}
    \label{fig:sim_real_setup}
\end{figure}
\subsubsection{Simulation Environment}
PyBullet is used to simulate the physics including realistic contacts, friction, and dynamics. The simulation environment (Fig. \ref{fig:sim_real_setup} left) consists of a dual-arm setup, the target object, and a vertical wall as the support object. One arm is used for pregrasping using a spheric end-effector, while the other arm is gripping using the default two parallel gripper end-effector. 

For training a robust policy, we randomly sample different values for the physical quantities described in Section \ref{sec:sim2real}.

The ADAM optimiser \cite{kingma2014adam} is used for optimising both the actor and critic parameters. The learning rate is set to $10^{-3}$, and the batch size is fixed at 100 samples. Both the actor and the critic contain 2 hidden layers, 64 neurons per layer. For SAC, we use a discount factor of  $\gamma = 0.99$, polyak averaging of $\theta_{\text{polyak}} = 0.995$, and optimise over the temperature parameter $\alpha$.
The policy is trained on i7-8700K without GPU, and converges towards a robust pregrasp policy after 250,000 samples which takes around one hour.

\subsection{Real-World Setup}

\begin{figure}[t]
\centering
\includegraphics[width=0.44\textwidth]{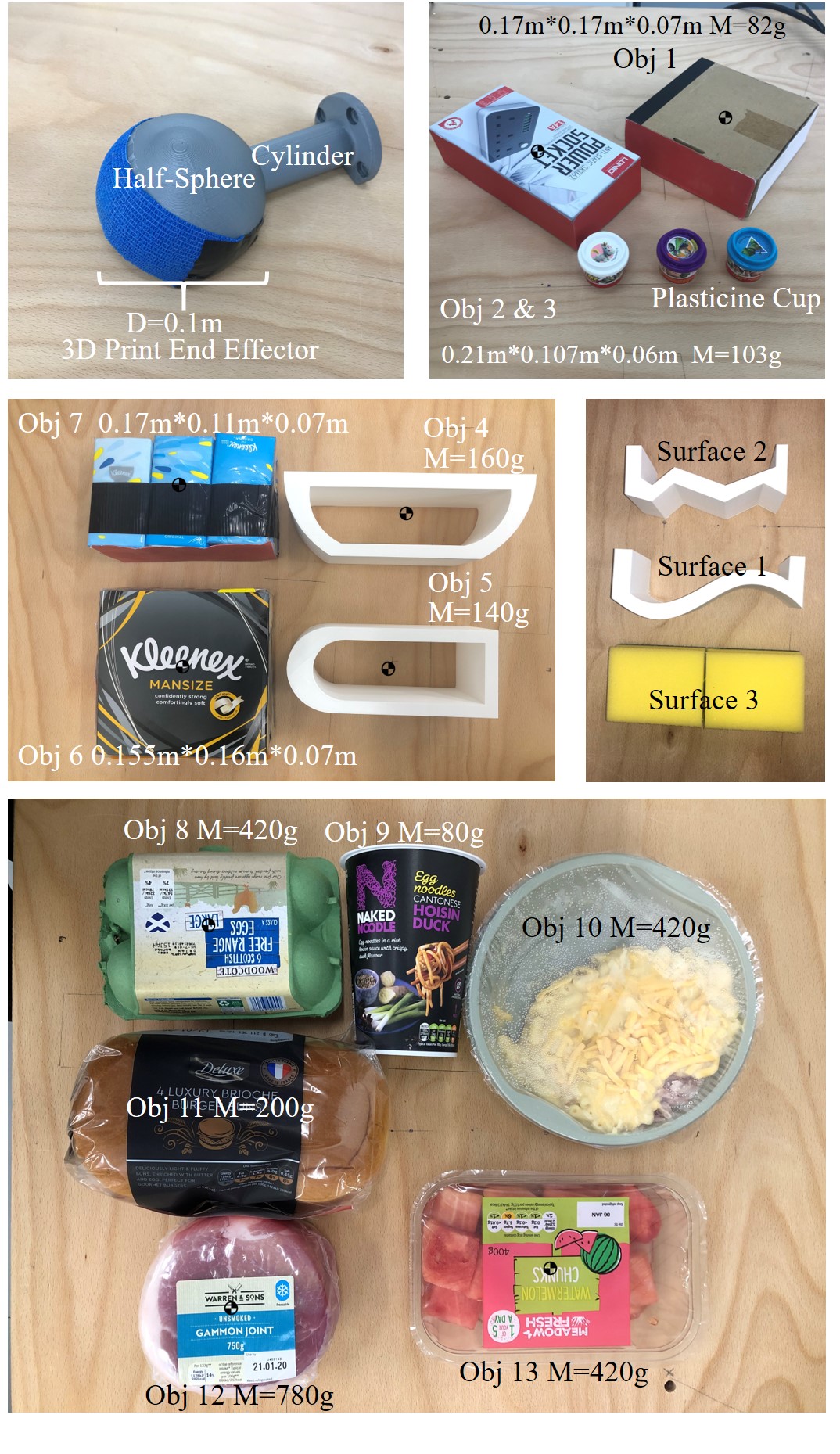}
\caption{End-effector, target objects, and support surfaces. Top left: 3D Printed end-effector with blue bandage to increase friction. Top right: object 1, 2, 3 with varying size and mass; plasticine cups for changing the objects' mass and as disturbances. Middle left: object 4, 5, 6, 7 as target object. Middle right: support surfaces 1 \& 2 with curve shape and zigzag shape respectively; sponges as deformable support Surface 3. Bottom: a box of eggs (Obj 8), a cup (Obj 9), a filled bowl (Obj 10), a bag of bread (Obj 11), a cylindric-shaped ham (Obj 12), a full plastic container (Obj 13).}
\label{fig:experimentSettingsDisplay}
\end{figure}

The experiments were conducted on the platform Panda Robotic Arms developed by Franka Emika with 7 Degrees of Freedom (DoF). The relative positions between robots, the target object, the support object are the same as the setup in simulation.
The robot's end-effector (Fig. \ref{fig:experimentSettingsDisplay} top left) operates on the centre vertical plane of the workspace, observed by an RGB camera. Through digital image processing, we extract the target object's position and pitch angle from the image. We first transfer RGB space into HSV space and use an upper and lower bound to segment the side which has already been pasted with a piece of red paper. These two bounds need to be determined first by humans. Then we use OpenCV library \cite{bradski2008learning} to extract a rectangle bounding box from this segmented image, which contains the necessary state information.

From joint encoder measurements, we further estimate the end effector's Cartesian $y$, $z$ position, pitch orientation, and the distance between the end effector and the centre point of the front surface of the target object. 

\begin{table}[h]
\caption{Coefficient of Friction between the target object and support surfaces. The static and kinetic CoF are indicated as first and second value respectively. The static CoF $\mu_s$ and kinetic CoF $\mu_k$ between the end effector's bandage and target object are 1.7 and 1.4 respectively. Blank data indicate non-conducted experiments. Object ID's as defined in Fig. \ref{fig:experimentSettingsDisplay}.}
\centering
\begin{tabular}{p{1cm}|p{1.0cm} p{1.0cm}|p{1.0cm} p{1.0cm}}
\hline
Obj ID & \multicolumn{2}{c}{Wood and Obj}  &\multicolumn{2}{c}{Surface 1, 2 and Obj}  \\ \hline
&$\mu_s$&$\mu_k$&$\mu_s$&$\mu_k$\\ \hline
Obj 1&0.32& 0.23 &  &\\
Obj 2, 3&0.24&0.17 & & \\
Obj 4, 5&0.26&0.19 & & \\
Obj 6&0.29&0.23&0.24&0.18\\
Obj 7&0.25&0.20& & \\
Obj 8&0.38&0.29&0.29&0.24\\
Obj 9&0.26&0.21&0.14&0.12\\
Obj 10&0.29&0.23&0.19&0.14\\
Obj 11&0.40&0.36&0.24&0.20\\
Obj 12&0.41&0.36&0.31&0.26\\
Obj 13&0.33&0.24&0.28&0.23\\ \hline
\end{tabular}
\label{table:cof}
\end{table}

As shown in Fig. \ref{fig:experimentSettingsDisplay}, real experiments used and tested a variety of target objects: Cuboids (Object 1, 2, 3), two 3D Printed objects with different contact shapes (Object 4, 5) and a compressible wrapped tissue bag (Object 6, 7). It shall be noted that \textit{only} the size of Object 1 is used for training. Object 2 - 7 are used to test both the robustness and generalisation ability of the policy. The coefficients of friction (CoF) between two pieces of bandage, the objects, and the support surface are in Table \ref{table:cof}. More experimental trials on Object 8 - 13 showing the generalisability of the policy can be found in the accompanying video. 

\subsection{Evaluation Metric}
To show the effectiveness of our DRL framework for pregrasp and the resulting policy, we define three evaluation metrics: 1. Task Completion Evaluation which is most fundamental metric to test whether an object is successfully lifted up and can be grasped at a feasible angle. 2. Robustness Evaluation which includes two definition, robustness towards disturbances, and robustness to different initial conditions. 3. Generalisation Evaluation which evaluates the generalisation ability to different objects and different support surfaces.

\subsubsection{Task Completion Evaluation}
\begin{figure}[t]
    \centering
    \includegraphics[width=0.48\textwidth]{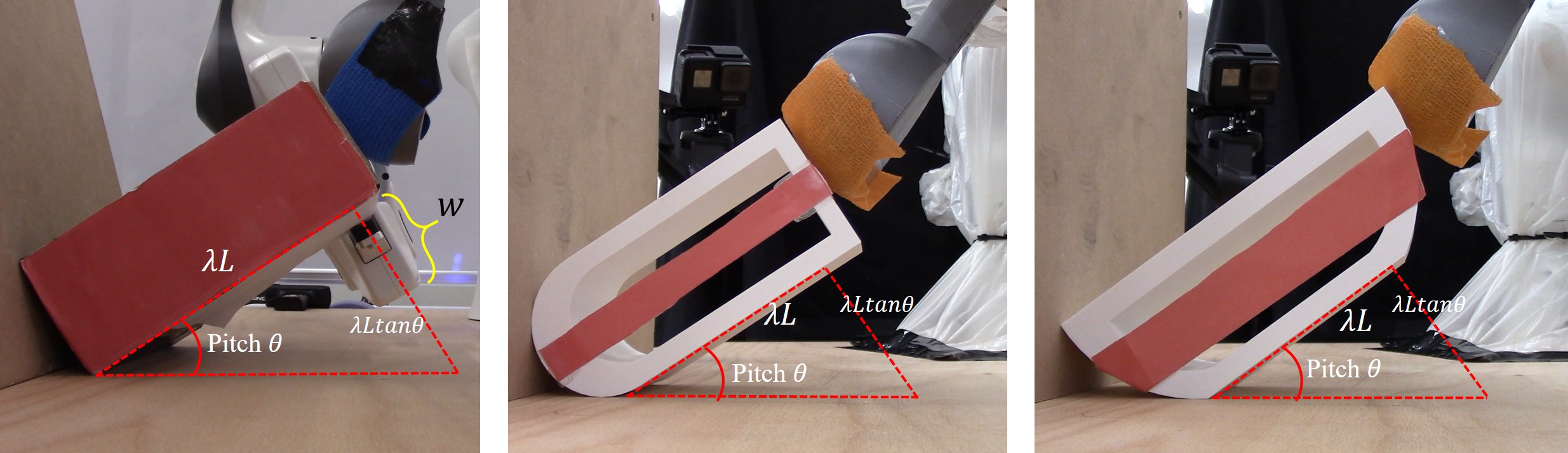}
    \caption{Task completion of 3 objects with varying shapes: all lifting angles were above a threshold for enough clearance.}
    \label{fig:completion}
\vspace{-2mm}
\end{figure}

For task completion, we propose to use the pregrasp lift angle as success indicator for whether pregrasp lifting is finished under different scenarios. For a success, the lift angle must be larger than a threshold such that the space underneath the target object is large enough for the gripper to grasp (Fig. \ref{fig:completion}):
\[ \lambda L \text{tan} \theta > w, \]
with length $L$ and pitch angle $\theta$ of the object, and $\lambda \in [0,1]$ indicating where the gripper's centre locates. We set $\lambda = 0.9$, and the gripper will target at around $90\%$ length to grasp. The threshold $w$ is defined to be $w = 0.05m$ in our experiment.

Furthermore, to indicate the quality of the pregrasp lift, we present the largest lift pitch angle in all evaluation tests.

\subsubsection{Robustness Evaluation}
Robustness evaluation includes two parts, robustness towards external disturbances, and robustness to different initial conditions.

To evaluate robustness against disturbances, we apply external forces on the target object. An object from a certain height will be dropped above the CoM of the object to find the max angular momentum that will make the target object fall. The angular momentum that the falling object applies to the target can be calculated by $L = r \times mv = mr\sqrt{2gh}$, with height $h$ and mass $m$ of the falling object, moment arm $r$ considering that grasp manipulation might cause a external torque on the object after lifting.

To evaluate robustness to different initial conditions, the object is placed in different $x, y$ positions and yaw poses (Fig. \ref{fig:robustness}). A robust agent should achieve success on a wide range of initial positions and poses. 

\subsubsection{Generalisation Evaluation}
Generalisation ability is measured by how many differences can the policy tolerate, which can be the discrepancy between simulation and real-world physics properties, and differences among experiments settings. We vary properties of the target, such as object size, shape, surface friction coefficient and mass. Furthermore, we test the generalisation ability by changing the type of support surfaces from vertical surfaces to inclined, deformable objects, and objects with different shapes.

\section{Experiments} 
This section presents the experiments and analysis towards the robustness and generalisation ability of our method. We visualize the learned actions as vector fields to provide an intuitive interpretation of the policy.

\subsection{Robustness Test}

\begin{figure}[t]
\vspace{-3mm}
    \includegraphics[width=0.48\textwidth]{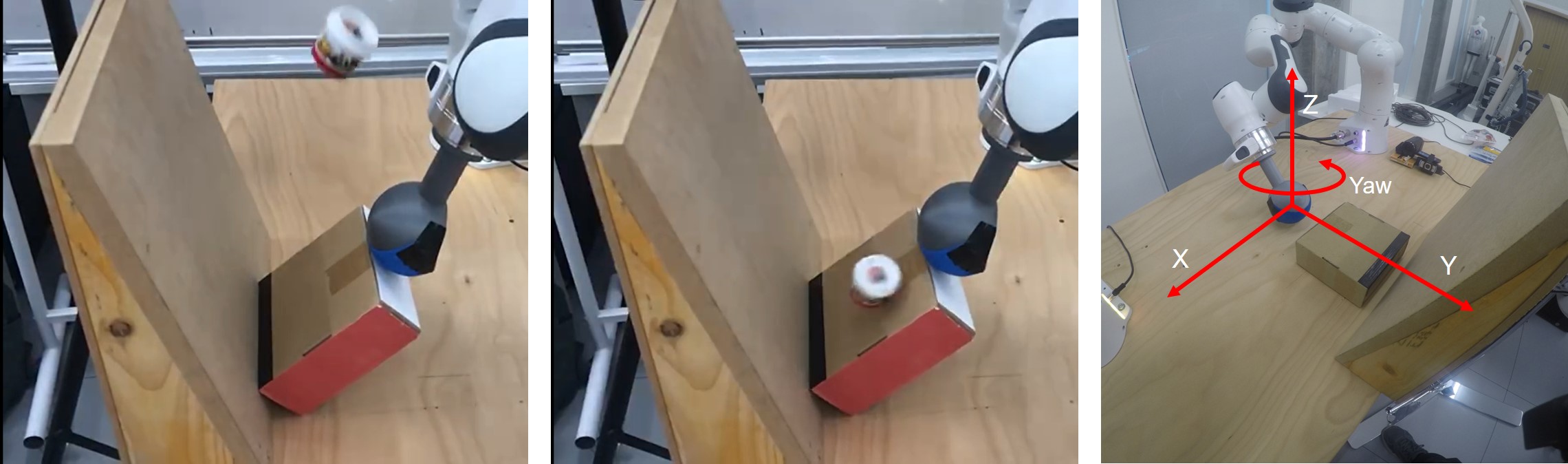}
    \caption{Two examples of testing robustness. {Left: }An object dropped from 0.5m height. {Middle: }The object collides with the target object by an impulse disturbance. {Right: }Objects' initial positions ($x, y$) and poses (yaw).}
    \label{fig:robustness}
\vspace{-5mm}
\end{figure}

In real-world tests, we use falling objects (Fig. \ref{fig:robustness}) to test the robustness of this lifting action. We choose plasticine cups (Fig. \ref{fig:experimentSettingsDisplay}) as our test unit, which weighs 0.028 kg (1 ounce) per cup. The falling height is 0.5 m above the collision point, and the corresponding angular momentum is 0.014 $kg m^2/s$.
The target object itself weights 0.082 kg. The policy robustly withstands an angular momentum of 0.028 $kg m^2/s$ equating to two plasticine cups, more than half of the target objects weight, being dropped simultaneously.

To evaluate on different initial positions and poses (Fig. \ref{fig:robustness}), we use Object 6. The $y$ position ranges from 0.05 m to 0.26 m, the longest distance from the object to the support surface is 0.21 m which is 1.3 times its length, and the shortest distance is 0 m where the object is next to the surface initially. The position on X ranges from -0.05 m to 0.05 m, from 30\% on the left side to 70\% to the right. The pose on Yaw angles from -0.3 rad to 0.3 rad. See results in Fig. \ref{fig:robustness and generalisation}.

\subsection{Generalisation Test}
The policy was trained only with Object 1 (Fig. \ref{fig:experimentSettingsDisplay}) and a vertical wall as the support surface. To test the generalisation, we varied both the parameters of the object and the support surface in the experiments. Furthermore, we tested the policy's ability to pregrasp under varying environmental conditions, and introduced variations in the contact friction between the object and the support surface, object size, surface inclination, and different support surfaces.

\subsubsection{Varying Object Parameters}

\begin{figure}[t]
    \centering
    \begin{subfigure}{0.46\textwidth}
        \centering
        \includegraphics[ width=0.48\textwidth]{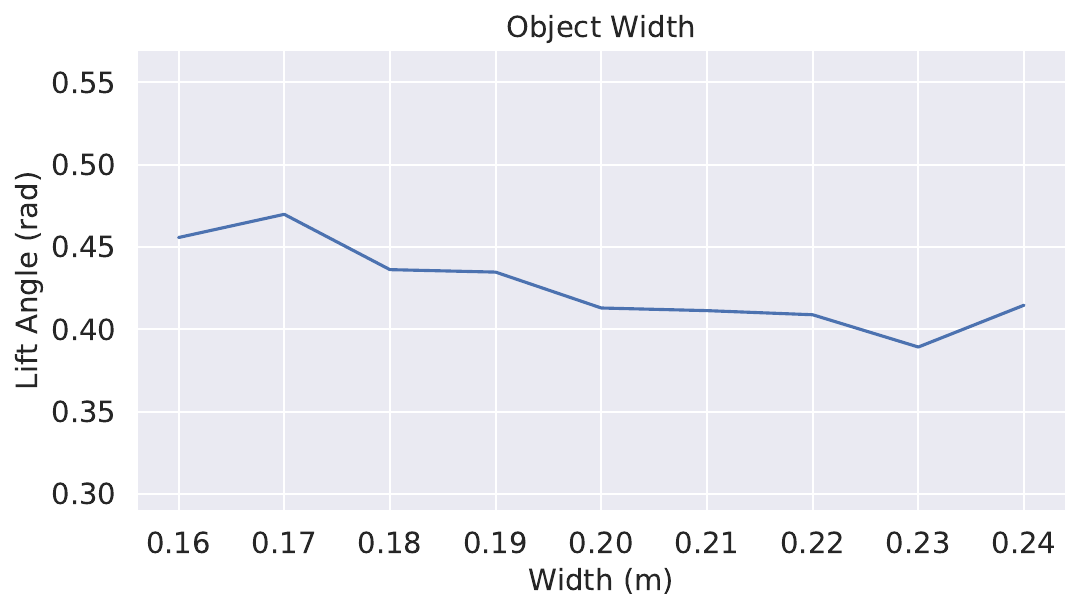}
        \includegraphics[ width=0.48\textwidth]{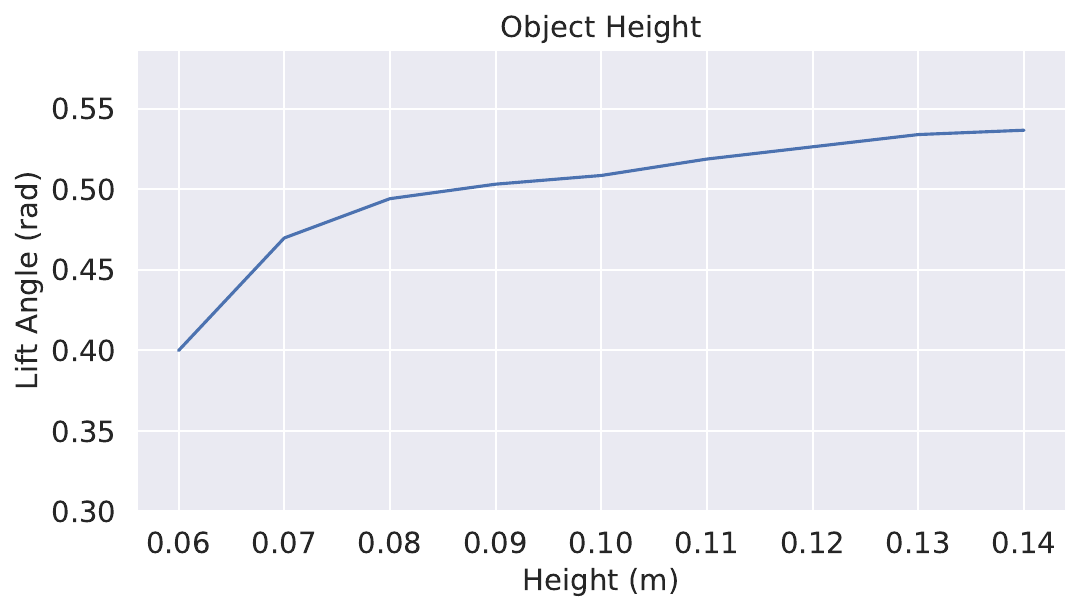}
    \end{subfigure}
    \begin{subfigure}{0.46\textwidth}
        \centering
        \includegraphics[ width=0.48\textwidth]{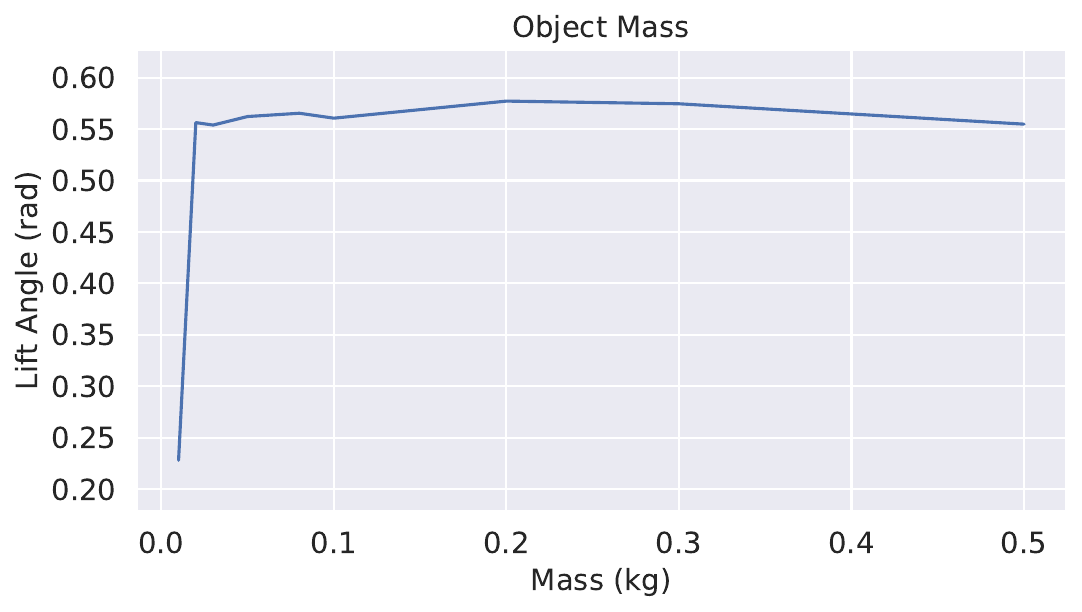}
        \includegraphics[ width=0.48\textwidth]{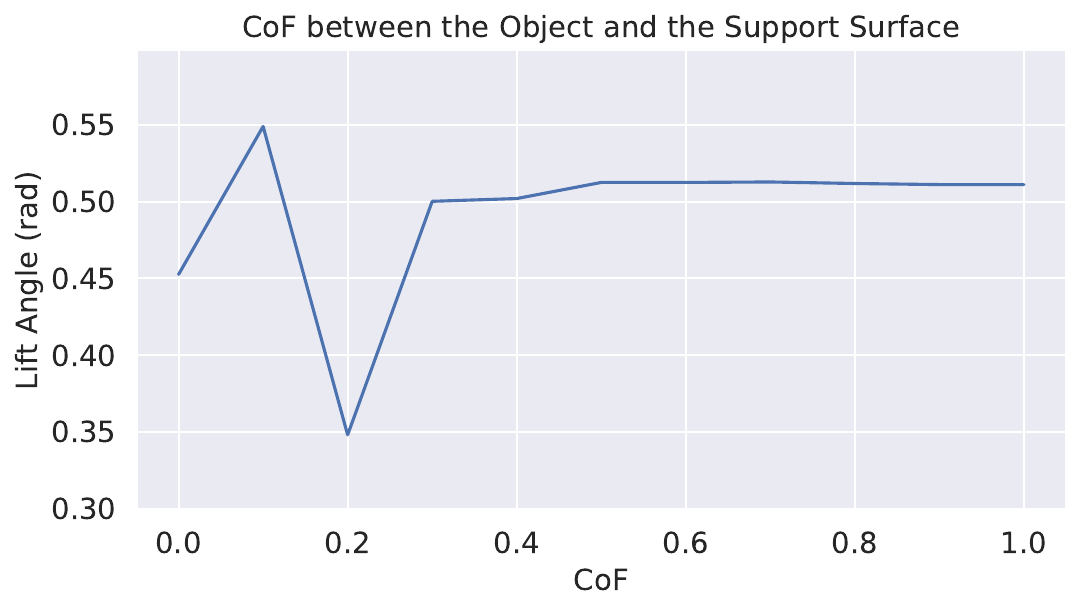}
    \end{subfigure}
    \caption{Robustness and generalisation tests in simulation (success when lift angle $> 0.32$ rad). Variation in the width (top left) and height (top right), the mass (bottom left) of the target object, and varying the coefficient of friction (bottom right) between target object and support surface.}
    \label{fig:object_param}
    \vspace{-5mm}
\end{figure}

We change the parameters in simulation, including the mass, the coefficient of friction, and size (height, width and length) (Fig. \ref{fig:object_param}): 

\textit{Coefficient of Friction:} For real-world experiments, we use different boxes with different contact surface material. Their corresponding CoF is shown in Table \ref{table:cof}.
 
\begin{figure}[t]
    \centering
    \begin{subfigure}{0.46\textwidth}
        \centering
        \includegraphics[ width=0.48\textwidth]{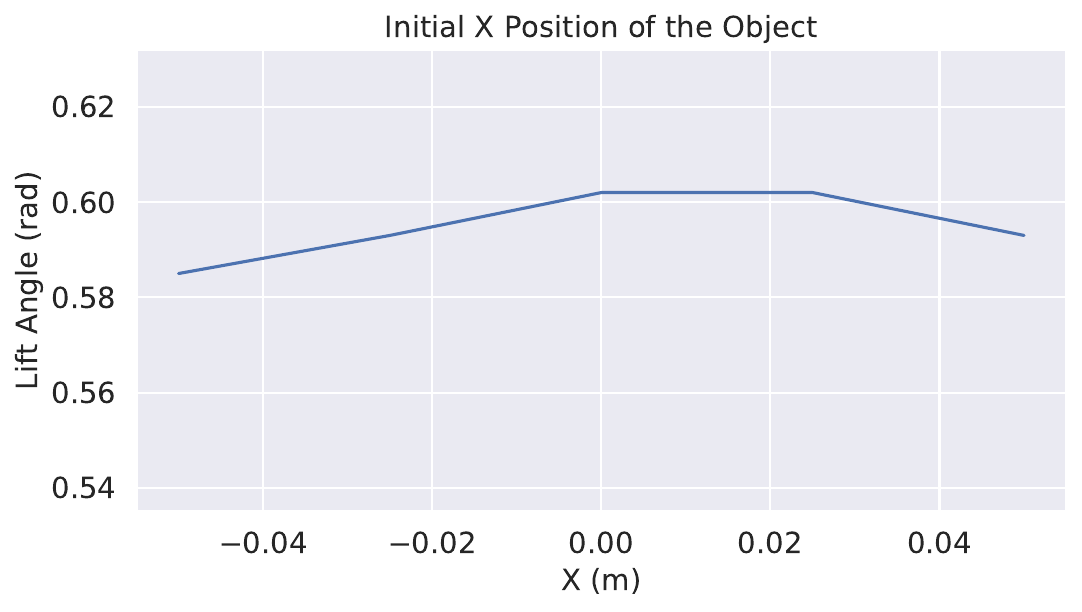}
        \includegraphics[ width=0.48\textwidth]{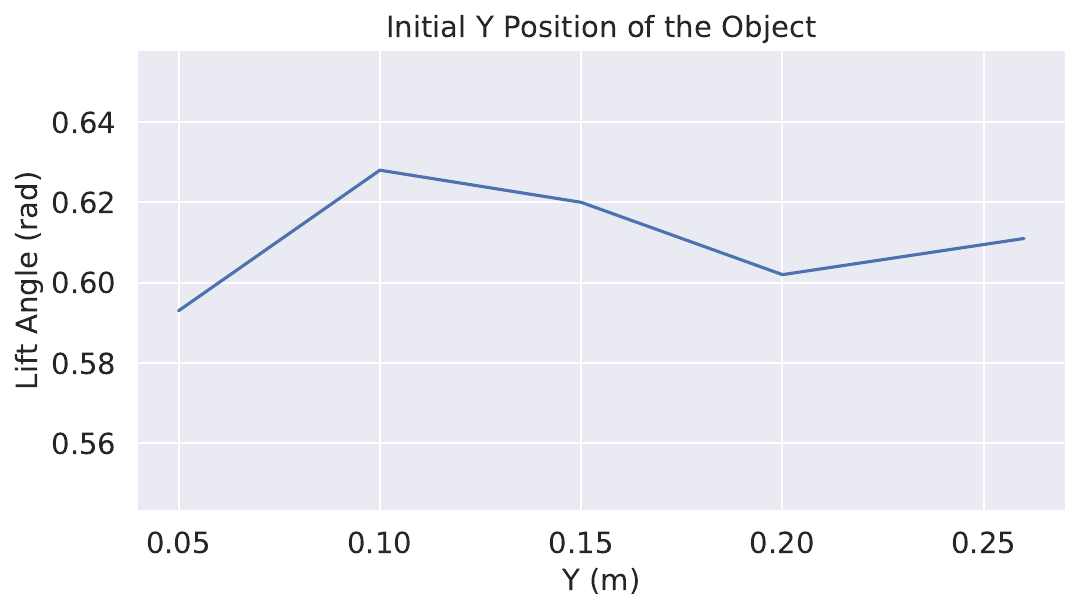}
    \end{subfigure}
    \begin{subfigure}{0.46\textwidth}
        \centering
        \includegraphics[ width=0.48\textwidth]{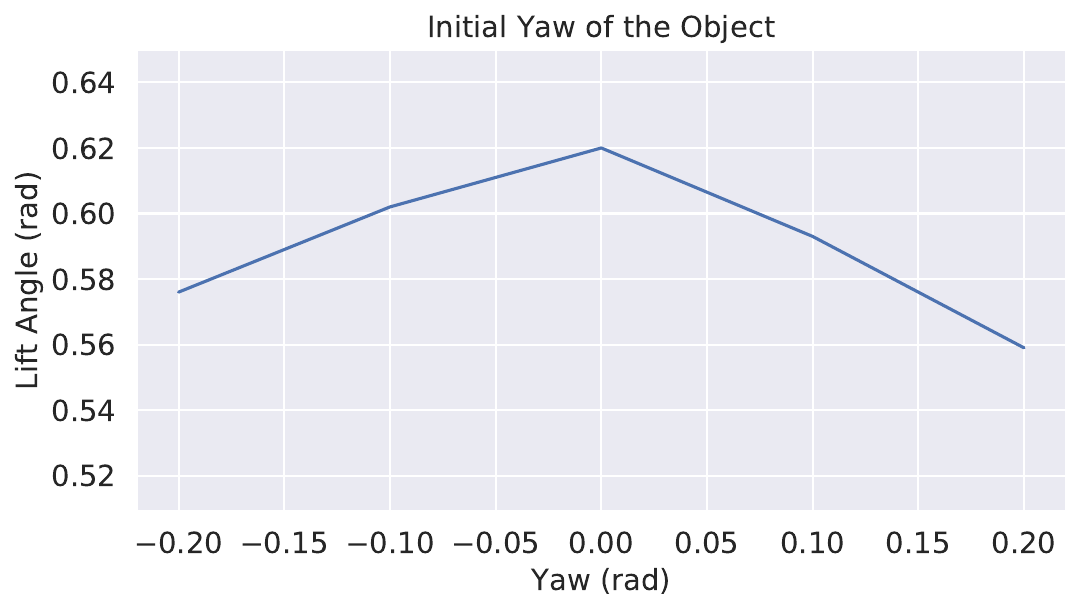}
        \includegraphics[ width=0.48\textwidth]{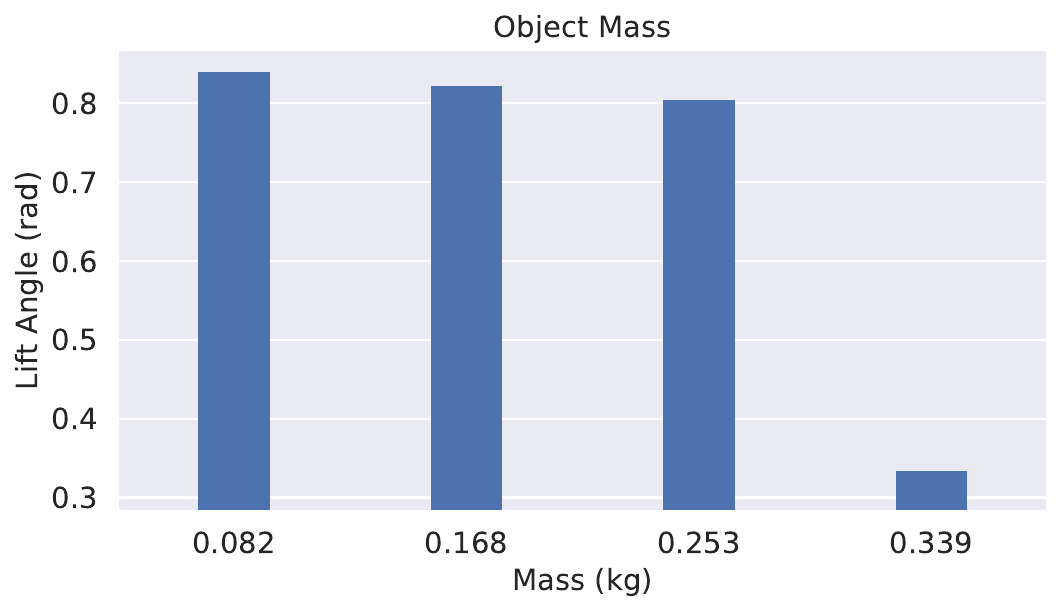}
    \end{subfigure}
    \caption{Robustness and generalisation tests in real experiments with different initial conditions.
        {Top left:} initial $x$ from -0.05 m to 0.05 m.
        {Top right:} initial $y$ from 0.0 m to 0.26 m. 
        {Bottom left:} initial yaw angle from -0.2 rad to 0.2 rad.
        {Bottom right:} objects with different mass.
}
    \label{fig:robustness and generalisation}
\vspace{-3mm}
\end{figure}

\textit{Weight:} The target object used in simulation training weighs 0.02kg. In real-world experiments, we range this mass from 0.08kg to 0.23kg by filling plasticine cups ($0.028kg$ per cup) into the box. Additionally to a change of mass, the CoM of the target object will dynamically shift due to randomly moving cups inside the box. The result is shown in Fig. \ref{fig:robustness and generalisation} bottom right.

\textit{Size:} In simulation we adjust width and height individually to obtain the graph between lift angles over width and height. In real-world test, we use Object 1 - 7 which have different sizes to show our agent's generalisation ability in Table \ref{tbl:realSize} and in Fig. \ref{fig:wall_obj}.

\textit{Shape:} In reality, we use Object 4 and 5 to evaluate our generalisation ability on different shapes. Object 4 has bowl-shape, and Object 5's contact shape is a half-circle.

\textit{Stiffness:} Additional to the rigid objects, a wrapped tissue box (Object 7), as a deformable object which can be compressed during lifting, was used in Fig. \ref{fig:wall_obj}. Three contact parts will deform during pregrasp: The part between object and support surface, the part between object and end effector, and the part between object and table.

\begin{table}[h]
\caption{Task Completion Evaluation on Objects with Different Sizes}
\centering
\begin{tabular}{llllllll}
\hline
Object ID             & 1 & 2 & 3 & 4 & 5 & 6 & 7 \\ \hline
Lift Angle (rad)  & .838 & .314 & .611 & .602 & .585 & .622 & .593 \\
Threshold(rad) & .316 & .259 & .479 & .378 & .404 & .334 & .316 \\ \hline
\end{tabular}
\label{tbl:realSize}
\end{table}


\subsubsection{Varying Support Surfaces}

\label{sec:support}
\begin{figure}[t]
    \centering
    \includegraphics[width=70mm]{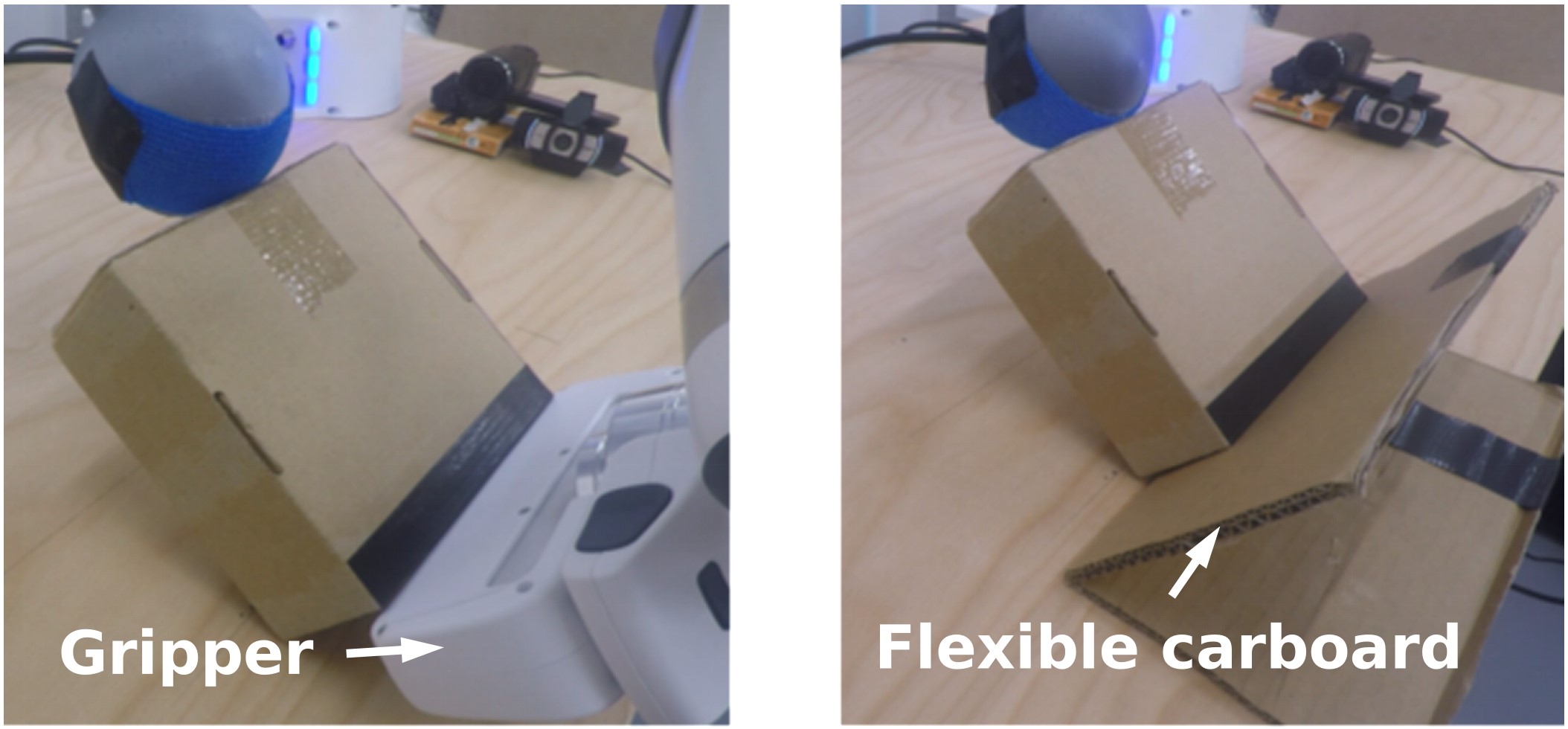}
    \caption{Two support surfaces with different shapes and firmness in our experiments: the gripper fingertips (left) and a flexible, deformable cardboard (right).}
    \label{fig:wall}
\vspace{-5mm}
\end{figure}

\begin{figure}[t]
    \centering
    \includegraphics[width=0.48\textwidth]{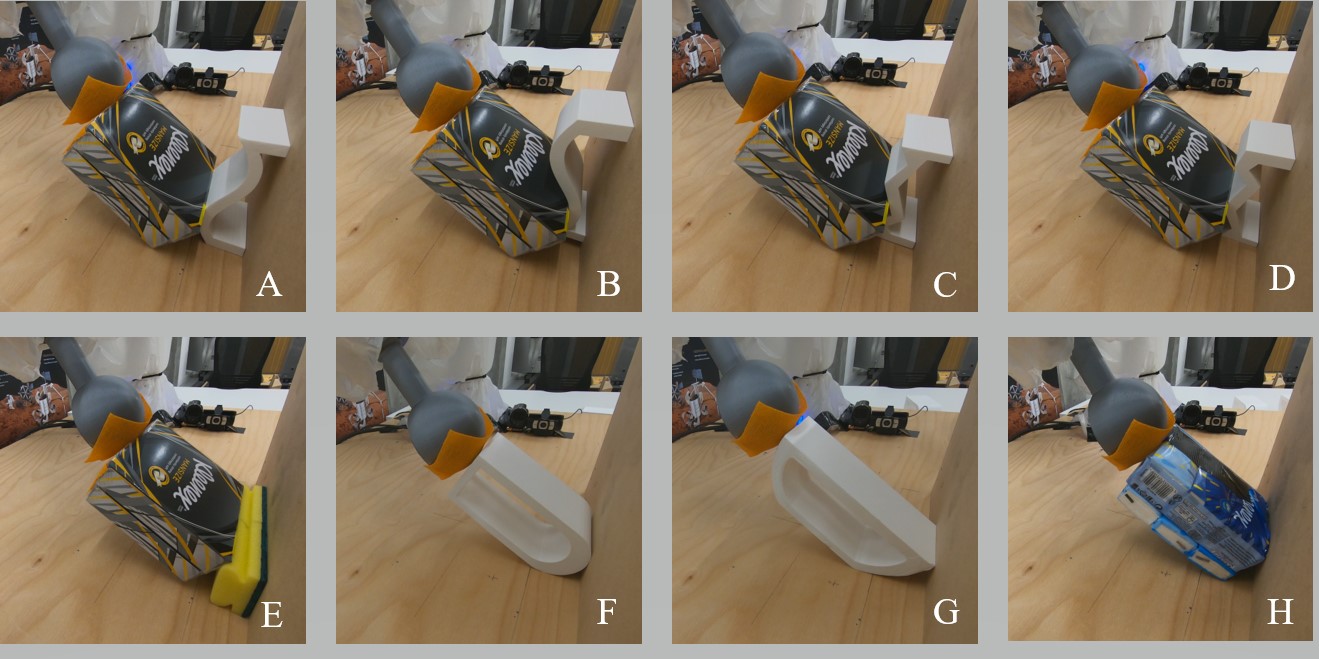}
    \caption{Combinations of different target objects and support surfaces.
    {A:} Surface 1 \& Obj 6.
    {B:} Surface 1 (upside down) \& Obj 6.
    {C:} Surface 2 \& Obj 6.
    {D:} Surface 2 (upside down) \& Obj 6.
    {E:} Surface 3 \& Obj 6.
    {F:} Wooden Wall \& Obj 5.
    {G:} Wooden Wall \& Obj 4.
    {H:} Wooden Wall \& Obj 7.
     }
    \label{fig:wall_obj}
\end{figure}

Although the policy is trained using a vertical wall as support surface, the policy learned by the agent can generalize and operate on different support surfaces (Fig. \ref{fig:wall}, Fig. \ref{fig:wall_obj}), such as various inclined surfaces with different inclinations, deformable surface made out of thin cardboard or a piece of sponge, the fingertips of the Grasper robot arm. This is because our method does not pose assumptions on the support surface properties. The policy can successfully generate pregrasp configurations in the following settings:

\textit{Fixed Inclination: }In the real-world we use three settings: $90^{\circ}$, $80^{\circ}$, $75^{\circ}$, and the lift angle of Object 1 is 0.803 rad, 0.820 rad, and 0.838 rad for each inclination angle.

\textit{Deformable Support Surface: } In this part, deformable support objects are used. We use a piece of cardboard to build a flexible surface (Fig. \ref{fig:wall}), which changes inclination angles when applied with larger force. We demonstrate that our agent trained on a vertical wall in simulation environment can generalize to this setting. In this case, the object is lifted to 0.803 rad which is larger than the threshold of 0.316 rad, and can be regarded as a feasible pose. For the deformable support surface, we use a piece of sponge which can also be compressed and deform during lifting. The lift angle is 0.611 rad where the threshold is 0.316 rad.

\textit{Support Surface with Different Shapes: } The support object in this experiment is two gripper fingertips (Fig. \ref{fig:wall}), but can be any fixed support surface, such as the base of the manipulator. The object is successfully lifted to 0.768 rad which is above the threshold of 0.316 rad. For other shapes, we use two 3D Printed surfaces, Surface 1 with the curved shape and Surface 2 with the zigzag shape.

\subsection{Policy Visualization and Analysis}
\begin{figure}[t]
\centering
\includegraphics[width=0.48\textwidth]{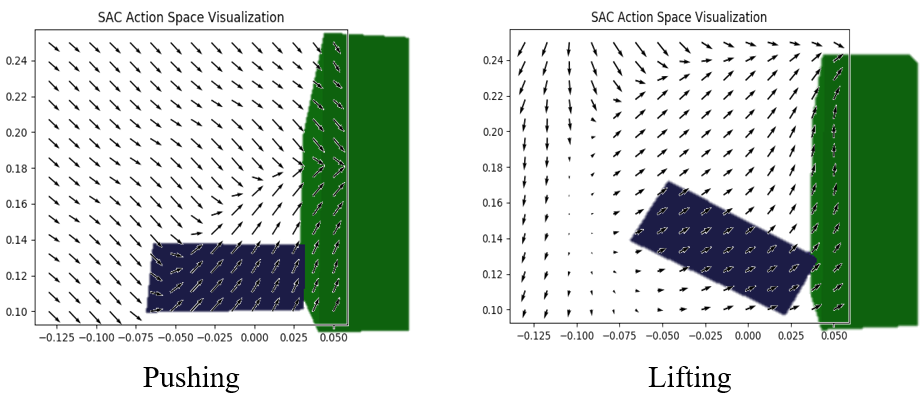}
\caption{End effector actions as vector fields in the Cartesian space. Blue rectangle is the target object, and green rectangle is the support surface.}
\label{fig:sacVisual}
\vspace{-5mm}
\end{figure} 
To understand better the fundamental mechanism of the learned policy, we conducted an analysis and visualization of the trained network's action space. In Fig. \ref{fig:sacVisual}, we can seen that the policy learns a mapping from spatial positions to actions, where the pitch angle of the end effector is fixed in this visualization to provide a 2D representation for $y$ and $z$ directions. We find that the policy learns to formulate actions as a field of attraction or a vector field, resulting in attractions for each state in the continuous space. During pushing and lifting, the end effector in the space will be guided based on the state feedback, and thus converge to the vector field automatically for following the task trajectory, which is robust to environmental disturbances. 

\section{Conclusions}
In this paper, we propose a Deep Reinforcement Learning Framework using SAC to control the robot arm, learning to push and lift those flat thin objects in ungraspable poses on the table. Compared to previous methods, we use less sensors and deal with more complicated scenarios. Besides the framework, with the proper state and action space representation, our SAC policy trained in simulation environment can directly transfer into real-world experiments with dynamic initialisation and reference initialisation. To better compare and evaluate the performance of the SAC policy, we further propose three metrics, including task completion, robustness and generalisation ability. Our extensive experiments in both simulation environment and real-world demonstrate the success of our framework. 

\addtolength{\textheight}{-12cm}   

\bibliographystyle{IEEEtran}


\end{document}